\documentclass{edm_article}
\usepackage{algorithm}
\usepackage{algorithmic}
\begin{document}

\title{Enhancing Explainability of Knowledge Learning Paths: Causal Knowledge Networks}

\numberofauthors{4}
\author{
\alignauthor
\parbox{2in}{\centering
Yuang Wei\titlenote{Corresponding Author.\\
*Received by EDM (Educational Data Mining) 2024, Human-Centric eXplainable AI in Education.\\
*Received 28 May 2024; Received in revised form 19 June 2024; Accepted 29 June 2024.}\\
       \affaddr{Lab of Artificial Intelligence for Education}\\
       \affaddr{East China Normal University}\\
       \affaddr{Shanghai, China}\\
       \email{philrain@foxmail.com}}
\alignauthor
\parbox{2in}{\centering
Yizhou Zhou\\
       \affaddr{School of Design and Engineering}\\
       \affaddr{National University of Singapore}\\
       \affaddr{Singapore}\\
       \email{e1010686@u.nus.edu}}
\and
\alignauthor
\parbox{2in}{\centering
Yuan-Hao Jiang\\
       \affaddr{Lab of Artificial Intelligence for Education}\\
       \affaddr{East China Normal University}\\
       \affaddr{Shanghai, China}\\
       \email{yuanhao.cs.edu@gmail.com}}
\alignauthor
\parbox{2in}{\centering
Bo Jiang\\
       \affaddr{Lab of Artificial Intelligence for Education}\\
       \affaddr{East China Normal University}\\
       \affaddr{Shanghai, China}\\
       \email{bjiang@deit.ecnu.edu.cn}}
}

\maketitle

\begin{abstract}
 A reliable knowledge structure is a prerequisite for building effective adaptive learning systems and intelligent tutoring systems. Pursuing an explainable and trustworthy knowledge structure, we propose a method for constructing causal knowledge networks. This approach leverages Bayesian networks as a foundation and incorporates causal relationship analysis to derive a causal network. Additionally, we introduce a dependable knowledge-learning path recommendation technique built upon this framework, improving teaching and learning quality while maintaining transparency in the decision-making process.
\end{abstract}


\section{Introduction}
The interconnected knowledge system, comprised of subject knowledge components, forms the basis of Intelligent Tutoring Systems (ITS)\cite{c1}. As the fundamental unit in teaching activities, the process of teaching and learning new knowledge usually follows a sequential methodology based on predefined teaching objectives\cite{c2}. As a result, the relationships and learning sequences among knowledge components within the system greatly influence learner outcomes. Additionally, these relationships can be utilized for domain knowledge modeling, learning recommendations, and even the construction of knowledge graphs\cite{al2015ontologies,grivokostopoulou2019ontology,sarwar2019ontology}. Such graphs incorporate emerged knowledge, target knowledge, and relationships throughout the learning process, generating multiple learning paths and facilitating path recommendations\cite{zhu2018multi}. 
Currently, most studies on knowledge component relationships are focused on correlations rather than causations. Correlations lack true explainability, as exemplified by the saying ``Storks Deliver Babies"\cite{matthews2000storks}, which illustrates correlation but not causation, thus failing to prove explainability. In the field of deep learning, correlation discovery has been extensively studied, yielding many excellent models. However, the demand for explainability in education renders most deep "black box" models insufficient. Examples include graph structure learning based on Graph Neural Networks\cite{jin2020graph} and unsupervised deep graph structure learning\cite{liu2022towards}. Traditional network structure learning methods, such as Bayesian network structure learning, although explainable, often struggle to accurately identify causal structures. Therefore, finding the most accurate causal relationships while maintaining explainability is key to completing knowledge component network structure learning.

This research aims to explore and understand the relationships between knowledge components, focusing on the nature and causes of these relationships. As previously discussed, correlation does not necessarily imply causation. Thus, relationships identified solely from data cannot be directly defined as causal, as this could be misleading. The study will focus on discovering causal relationships between each pair of knowledge components\cite{jin_can_2024,cox_causal_2023,enow_investigating_2023}. As Plato (1961) suggests, contemplating relationships should involve seeking the causes of each thing: why it comes into existence, why it ceases to exist, and why it exists in the first place. The study of causality spans various disciplines, all focused on addressing the fundamental question of ``Why?"\cite{nogueira2022methods} 

Therefore, it is essential to investigate an explainable and trustworthy knowledge relationship structure to answer ``Why" and provide a method of ``How" to utilize this structure. This model aims to enhance transparency in knowledge teaching and improve overall educational quality.

In this paper, we introduce the concept of a causal knowledge network and address some potential challenges. We begin by illustrating the construction of a causal knowledge network and then present a method for leveraging this structure for learning path recommendations, ultimately aiming to improve teaching quality\cite{paudel_transparency_2022,winkelmes_introduction_nodate}. Finally, we discuss the potential limitations and challenges associated with this approach.

\section{Related works}
Research on the relationships between knowledge components has predominantly focused on identifying prerequisite relationships, which are crucial for delineating the directional connections between concepts \cite{liang2018investigating}. Various data-driven approaches have been explored, including the development of preliminary tests and manual associations \cite{chen2015discovering, chen2016joint}. However, these strategies have shown limited effectiveness when dealing with large datasets or complex networks. While methods based on information theory and topic modeling offer better explainability, they require extensive manual intervention \cite{gordon2016modeling}. The rise of deep learning has advanced the application of semantic analysis in identifying prerequisite relationships, particularly in contexts like Wikipedia and MOOCs, though challenges in scalability and explainability remain \cite{pan2017prerequisite, liang2017recovering}.

Although these studies have successfully identified prerequisite relationships among knowledge components and considered them as a specific type of causal relationship to optimize teaching or learning sequences, uncovering latent causal relationships from student test data is more critical for adaptive learning systems. This is because learning outcome data can more accurately reflect students' mastery of content, aligning better with the essence of personalized learning \cite{carlson2022fundamental}. Causal relationships, including prerequisite ones, more authentically represent the connections between elements \cite{yu2022learning}. Therefore, there is a need for a causality-based discovery approach to extract the causal relationships between knowledge components from student test data.

\section{Causal network of knowledge components}

\begin{figure*}[tb]
\Description{Experiment design and data representation}  
\centering
\includegraphics[width=0.85\textwidth]{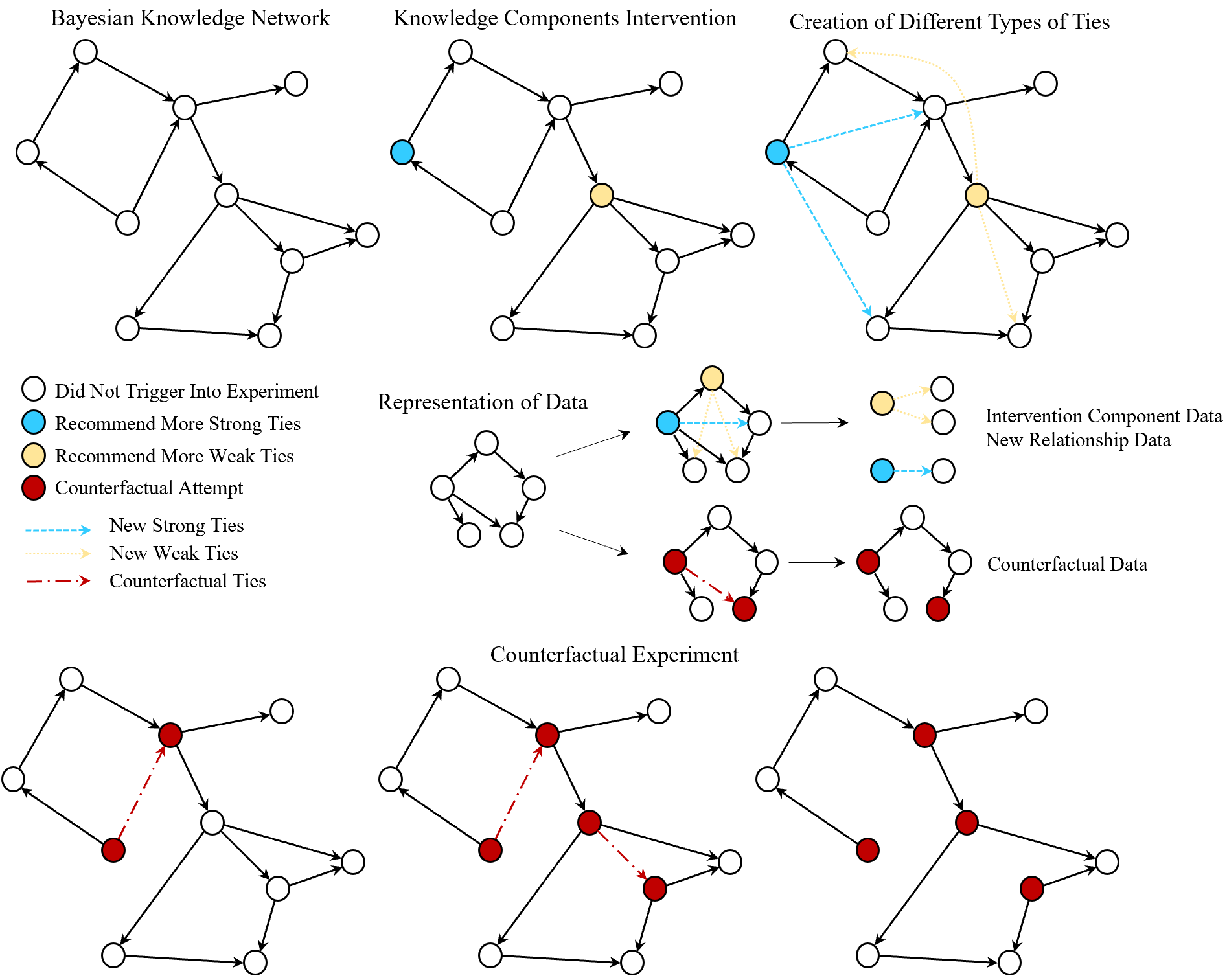}
\caption{Experiment design and data representation: This figure demonstrates the use of Bayesian Knowledge Networks to investigate causal relationships among knowledge components through targeted interventions and counterfactual experiments. The process involves recommending new strong and weak ties, representing the data, and conducting counterfactual experiments to validate the potential causal structures.}
\label{fig1}
\end{figure*}

To capture the causal relationships among learning concepts, we begin by establishing a foundational knowledge network using available data. Bayesian networks—probabilistic graphical models represented through Directed Acyclic Graphs (DAGs) and Conditional Probability Tables (CPTs)—are used to model the relationships between variables, making them ideal for constructing knowledge network structures due to their directed and probabilistic nature\cite{pinto2009using}. When calculating the Bayesian network structure, we use the Bayesian Information Criterion (BIC) \cite{vrieze2012model} as the scoring function. Since we will later update the network with causal effects, any function can initially be selected. So, to compute the structure of a Bayesian Network (BN) using the BIC score, we start by defining the BIC score for a given Bayesian Network structure \( G \) with parameter set \( \theta \) and dataset \( D \). The BIC score is given by:

\begin{equation}
    \text{BIC}(G, \theta, D) = \log P(D \mid G, \theta) - \frac{|\theta|}{2} \log N
\end{equation}
where \( P(D \mid G, \theta) \) is the likelihood of the data given the network structure and parameters, \( |\theta| \) is the number of parameters in the model, and \( N \) is the number of data points.

Next, we compute the likelihood for each node \( X_i \) in the network with parents \( \text{Pa}(X_i) \):

\begin{equation}
    P(D \mid G, \theta) = \prod_{i=1}^{n} \prod_{j=1}^{q_i} \prod_{k=1}^{r_i} \left( P(X_i = k \mid \text{Pa}(X_i) = j, \theta) \right)^{N_{ijk}}
\end{equation}
where \( n \) is the number of nodes, \( q_i \) is the number of parent configurations for node \( X_i \), \( r_i \) is the number of states of node \( X_i \), and \( N_{ijk} \) is the number of instances in the data where \( X_i = k \) and \( \text{Pa}(X_i) = j \).

The log-likelihood component of the BIC score is computed as follows:

\begin{equation}
\begin{aligned}
    &\log P(D \mid G, \theta) \\
    &= \sum_{i=1}^{n} \sum_{j=1}^{q_i} \sum_{k=1}^{r_i} N_{ijk} \log P(X_i = k \mid \text{Pa}(X_i) = j, \theta)
\end{aligned}
\end{equation}

To determine the number of parameters \( |\theta| \) in the network, we use the formula:
\begin{equation}
    |\theta| = \sum_{i=1}^{n} q_i(r_i - 1)
\end{equation}

Finally, we substitute the log-likelihood and the number of parameters into the BIC formula to calculate the BIC score:

\begin{equation}
\begin{aligned}
&\text{BIC}(G, \theta, D) = \left( \sum_{i=1}^{n} \sum_{j=1}^{q_i} \sum_{k=1}^{r_i} N_{ijk} \log P(X_i = k \mid \text{Pa}(X_i) = j) \right) \ \\
&- \frac{1}{2} \left( \sum_{i=1}^{n} (r_i - 1) q_i \right) \log N
\end{aligned}
\end{equation}

Then we can use a structure search algorithm to learn the network structure that can maximize the BIC function value. The algorithm structure is shown in Algorithm \ref{algo0}. The obtained network structure will be used as the initial structure for learning the causal network.

\begin{algorithm}
\caption{Structure Search Algorithm}
\label{algo0}
\begin{algorithmic}[1]
\STATE \textbf{Input:} Data set $D$, initial Bayesian Network structure $G_0$
\STATE \textbf{Output:} Optimal Bayesian Network structure $G^*$

\STATE Initialize $G \gets G_0$
\STATE Compute $\text{BIC}(G)$
\REPEAT
    \STATE improvement $\gets \text{false}$
    \STATE $G_{\text{best}} \gets G$
    \FOR{each possible modification $G'$ of $G$}
        \STATE Compute $\text{BIC}(G')$
        \IF{$\text{BIC}(G') > \text{BIC}(G_{\text{best}})$}
            \STATE $G_{\text{best}} \gets G'$
            \STATE improvement $\gets \text{true}$
        \ENDIF
    \ENDFOR
    \IF{improvement}
        \STATE $G \gets G_{\text{best}}$
    \ENDIF
\UNTIL{improvement is \text{false}}
\RETURN $G$
\end{algorithmic}
\end{algorithm}

Following the initial network construction, we delve into the task of uncovering the causal relationships between the knowledge components. This involves conducting interventions and counterfactual experiments on the basic network structure. We design intervention experiments to manipulate specific knowledge components using do-calculus, developed by Judea Pearl, enabling observation of changes in the probabilistic dependencies among the components. By doing so, we can assess whether the existing connections between concepts are robust and should be retained as strong links or if they should be weakened due to the intervention. These experiments are outlined in Fig. \ref{fig1}, which provides a visual representation of the fundamental operations involved.

Once we have intervened and observed the effects, we introduce new data to reevaluate the strength of the connections that have been formed or adjusted. This step is crucial for reinforcing our understanding of the causal relationships and for refining the network structure to better represent the true causal mechanisms at play.

In addition to intervention experiments, we also conduct counterfactual experiments, which involve hypothesizing alternative scenarios and examining how the network would respond under those conditions. During these experiments, we modify the node associations within the network to simulate the hypothesized conditions and then apply do-calculus\cite{pearl1995causal} to the altered network to explore the potential outcomes. This process allows us to test the robustness of the causal relationships under different hypothetical contexts.

The specific intervention process is as follows: first, we use the initially obtained network structure \( G_0 \) to represent the assumed causal relationships between variables. Let the variables in the network be \( X_1, X_2, ..., X_n \), and their causal relationships are represented by a directed acyclic graph (DAG) as \( G=(V, E) \), where \( V \) is the set of nodes and \( E \) is the set of directed edges.

To calculate the causal effect of node \( X_i \) on node \( X_j \), we can use Pearl's back-door criterion \cite{pearl2009causality}. The back-door criterion tells us that to calculate the causal effect of \( X_i \) on \( X_j \), we need to control for all non-descendant nodes of \( X_i \) and then intervene on \( X_i \). Therefore, we need to find the set \( Z \) of all non-descendant nodes belonging to \( X_i \), and for each node \( Z_k \) in \( Z \), calculate \( P(Z_k | pa(Z_k)) \), where \( pa(Z_k) \) is the parent node of \( Z_k \). Then, intervene on \( X_i \) by setting it to a specific value \( x_i' \), and calculate \( P(X_j | do(X_i=x_i'), Z) \). Finally, calculate the causal effect of \( X_i \) on \( X_j \), i.e., \( P(X_j | do(X_i=x_i')) \).

By combining insights from both intervention and counterfactual experiments, we construct a comprehensive causal network of knowledge components. This network not only reflects the probabilistic relationships between concepts but also provides a deeper understanding of the causal mechanisms that drive knowledge acquisition and mastery. The resulting causal network serves as a valuable tool for educators and learners alike, offering a detailed map of the interconnectedness of knowledge and a basis for targeted interventions to enhance learning outcomes.

\section{Knowledge learning path planning}

\begin{figure*}[tb]
\Description{Knowledge component learning path recommendation based on causal networks}  
\centering
\includegraphics[width=0.85\textwidth]{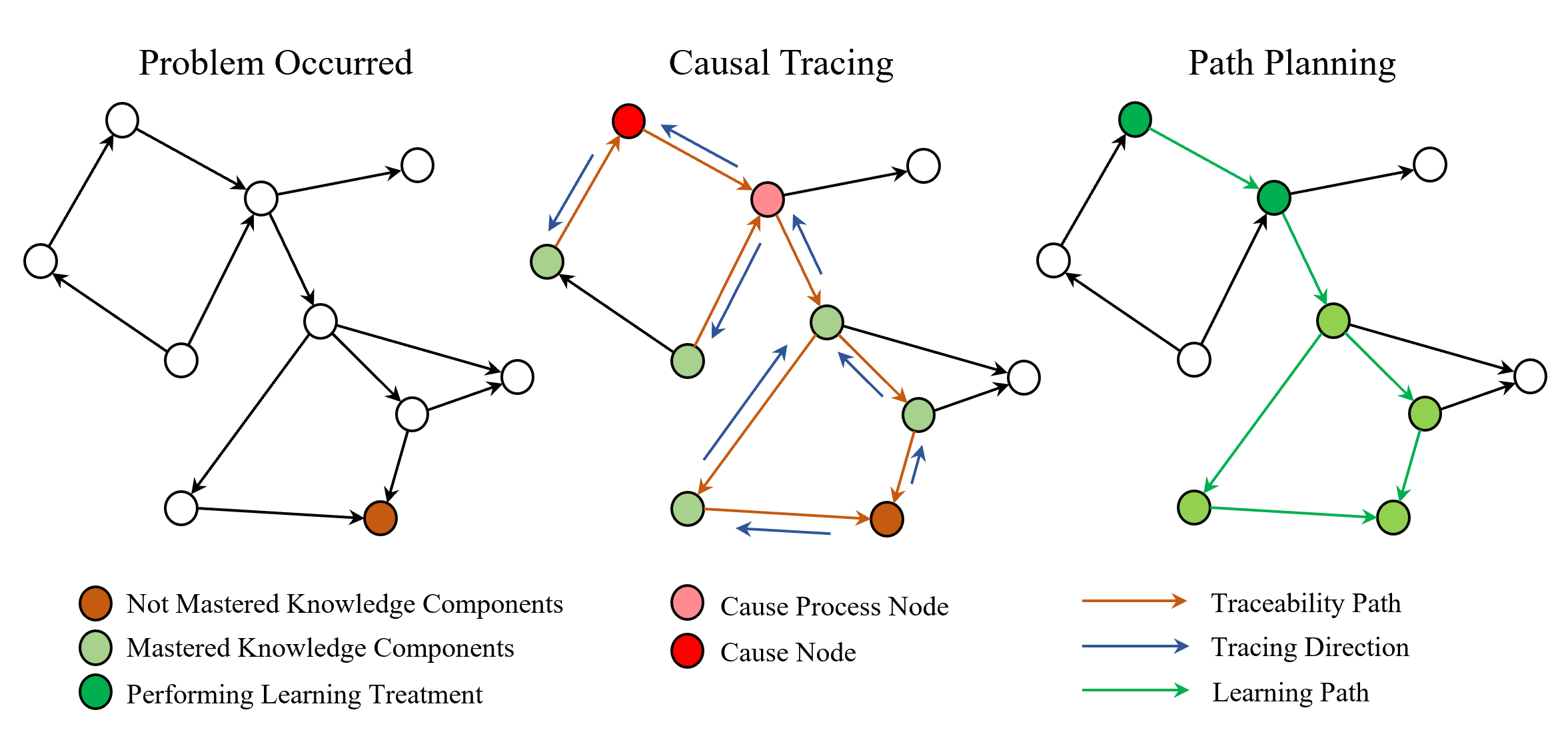}
\caption{Knowledge component learning path recommendation based on causal networks: The figure demonstrates the steps of addressing a learning problem in a knowledge network: identifying unmastered components, tracing the causal relationships of the issue, and planning an effective learning path to achieve mastery.} 
\label{fig2}
\end{figure*}
The causal network serves not merely as a static representation of knowledge interconnectivity but as a dynamic tool for educational planning. Each node within the network, fortified by the robustness of its causal linkages, becomes a critical checkpoint in an individual student's learning trajectory. 
 
To effectively assess and support learning, it is essential to first construct a comprehensive causal network. This network takes the form of a conceptual map that connects all relevant concepts and topics involved in the learning objectives through their causal relationships. Following this, an initial assessment of students' understanding is conducted through methods such as quizzes, interviews, or observations to determine their grasp of each knowledge component within the network. The assessment results will help us identify the knowledge components that students have not fully mastered, which are the gaps in their understanding. 

By analyzing the interconnections of these unmastered knowledge components within the causal network, we can trace the origins of the knowledge gaps and establish a path leading to the root nodes—the fundamental causes of the students' difficulties. Throughout this process, we highlight all problematic nodes, which are points where students are likely to encounter challenges and require additional support. After identifying the root nodes, we recommend that learners concentrate on strengthening their grasp of these critical concepts through additional practice, review sessions, or targeted instruction.

Ultimately, we provide learners with a detailed guide that serves as a roadmap for their systematic journey through the causal network, ensuring they address each concept logically and organized. This approach fills in knowledge gaps and builds a solid foundation of interconnected knowledge. 

To illustrate this process more clearly, we refer to Fig. \ref{fig2} in the text, which provides a visual representation of the causal network and the traced path. At the same time, to quickly find the path for students to solve the superficial problem step by step from the root problem, we build upon the network structure obtained in the previous section. By learning path tracing and based on the current mastery status of each knowledge component node, we identify the root problem node and then proceed to find the shortest path to the superficial problem node, obtaining the shortest learning path to facilitate student learning. The specific algorithmic process is illustrated in Algorithm \ref{AlgoFSP}.

\begin{algorithm}
\caption{Find Shortest Path in a Directed Graph}
\label{AlgoFSP}
\begin{algorithmic}[1]
\REQUIRE Directed graph $G$, source node $s$, target node $t$
\ENSURE Shortest path from $s$ to $t$ in $G$
\STATE Initialize an empty queue $Q$
\STATE Enqueue $s$ into $Q$
\STATE Initialize a dictionary $dist$ with all nodes in $G$ as keys, set $dist[s] = 0$ and $dist[v] = \infty$ for all other nodes $v$
\WHILE{$Q$ is not empty}
    \STATE Dequeue a node $u$ from $Q$
    \FOR{each neighbor $v$ of $u$ in $G$}
        \STATE Calculate $alt = dist[u] + \text{weight}(u, v)$
        \IF{$alt < dist[v]$}
            \STATE $dist[v] = alt$
            \STATE Enqueue $v$ into $Q$
        \ENDIF
    \ENDFOR
\ENDWHILE
\RETURN $dist[t]$
\end{algorithmic}
\end{algorithm}

\section{Experiment}
\subsection{Data preparation}

Before learning the causal network of knowledge components, it is necessary to collect data on students' actual learning processes. This type of data is similar to datasets like Assistment\footnote{https://sites.google.com/site/assistmentsdata/datasets} and Junyi\footnote{https://pslcdatashop.web.cmu.edu/DatasetInfo?datasetId=1198}. In our current experiments, we have collected data on the learning processes and outcomes of mathematics courses from 77 classes in 19 elementary schools and 7 middle schools across Shanghai, Sichuan, Jiangsu, and Beijing, China. These classes cover four grades, from fourth to seventh. Using common cognitive diagnostic methods such as knowledge tracing and the DINA model, we obtained students' mastery states of knowledge components. These mastery states, represented as time series data, serve as the foundation for constructing the causal knowledge network proposed in this study. The experimental data were obtained from our self-designed adaptive learning platform\footnote{http://web.ai-learning.cn/}, which served as the data foundation for constructing the network. 

Following is an example from a small-scale experiment. For larger-scale experiments and comparisons with other methods, please look forward to our future research publications.

\subsection{Experimental Example}
We demonstrate the construction process of a causal knowledge network through a small-scale experiment, with the algorithm workflow shown in Algorithm \ref{algo1}. 

The Algorithm \ref{algo1} begins with a learning performance dataset as input, which is then transformed into a knowledge mastery dataset containing student IDs and corresponding levels of knowledge proficiency. Subsequently, an initial knowledge network $D$ is constructed through correlation learning, and the following steps are iteratively executed while the student scores remain stable: the current knowledge network structure's score is calculated using the Bayesian Information Criterion (BIC), followed by the optimization of the network structure through Hill Climbing search to identify a better network structure $\rm{\textit{D}^{(new)}}$, which then updates $D$ to $\rm{\textit{D}^{(new)}}$. Once the optimization is complete, the algorithm returns a knowledge network $\rm{\textit{D}^{(Bayesian)}}$ that has been updated through Bayesian inference, and proceeds to traverse each edge of the network, using the Refute function to verify the causality of each edge. Ultimately, the algorithm outputs a knowledge network $\rm{\textit{D}^{(Causality)}}$ that has undergone causality analysis.

\begin{algorithm}[tb]
\caption{Constructing Causal Knowledge Networks}
\label{algo1}
  \begin{algorithmic}
    \STATE \textbf{Input:} Learning performance dataset = ${\rm{\{ id,}}{{\rm{s}}^{(id)}}{\rm{\} }}_{id = 1}^{student\_id}$
    \STATE Transform: Learning performance dataset $\Rightarrow$ Knowledge mastery dataset = ${\rm{\{ id,}}{{\rm{k}}^{(id)}}{\rm{\} }}_{id = 1}^{student\_id}$
    \STATE Initial knowledge network through correlation learning: $\rm{\textit{D}}$
    \STATE \textbf{while}\ ${\rm{ }}score$ is stable
    \STATE \quad$\rm{\textit{D}^{(new)}},\textit{score}$ = Algorithm 1(${\rm{\{ id,}}{{\rm{k}}^{(id)}}{\rm{\} }}_{id = 1}^{student\_id}$, $\rm{\textit{D}}$)
    \STATE \quad$\rm{\textit{D}}$ = $\rm{\textit{D}^{(new)}}$
    \STATE \textbf{return}\ $\rm{\textit{D}^{(Bayesian)}}$
    \STATE \textbf{for}\ ${\rm{ }}edge$ in $\rm{\textit{D}^{(Bayesian)}}$
    \STATE \quad$\rm{\textit{Causality}}$ = Refute(${\rm{\{ id,}}{{\rm{k}}^{(id)}}{\rm{\} }}_{id = 1}^{student\_id}$, ${\rm{ }}edge$)
    \STATE \textbf{Output}\ $\rm{\textit{D}^{(Causality)}}$
  \end{algorithmic}
\end{algorithm}

In the Algorithm \ref{algo1}, the Refute method is used to validate the reliability of the inferred causal relationships through interventions, which is the causal effect calculation method discussed previously. In this process, counterfactual data can be generated through sampling, perturbation, or other methods as needed for the experiment to verify the causal relationships. In the experimental example of this study, we calculate causal effects using only intervention methods and determine the strength of the causal relationships to decide whether to add or remove a particular edge. And \textit{BIC\_score} comes from \cite{schwarz1978estimating} and the hill-climbing algorithm from \cite{hillclimb}.

\begin{figure*}[tb]
\Description{Construction of knowledge component causal network}  
\centering
\includegraphics[width=0.85\textwidth]{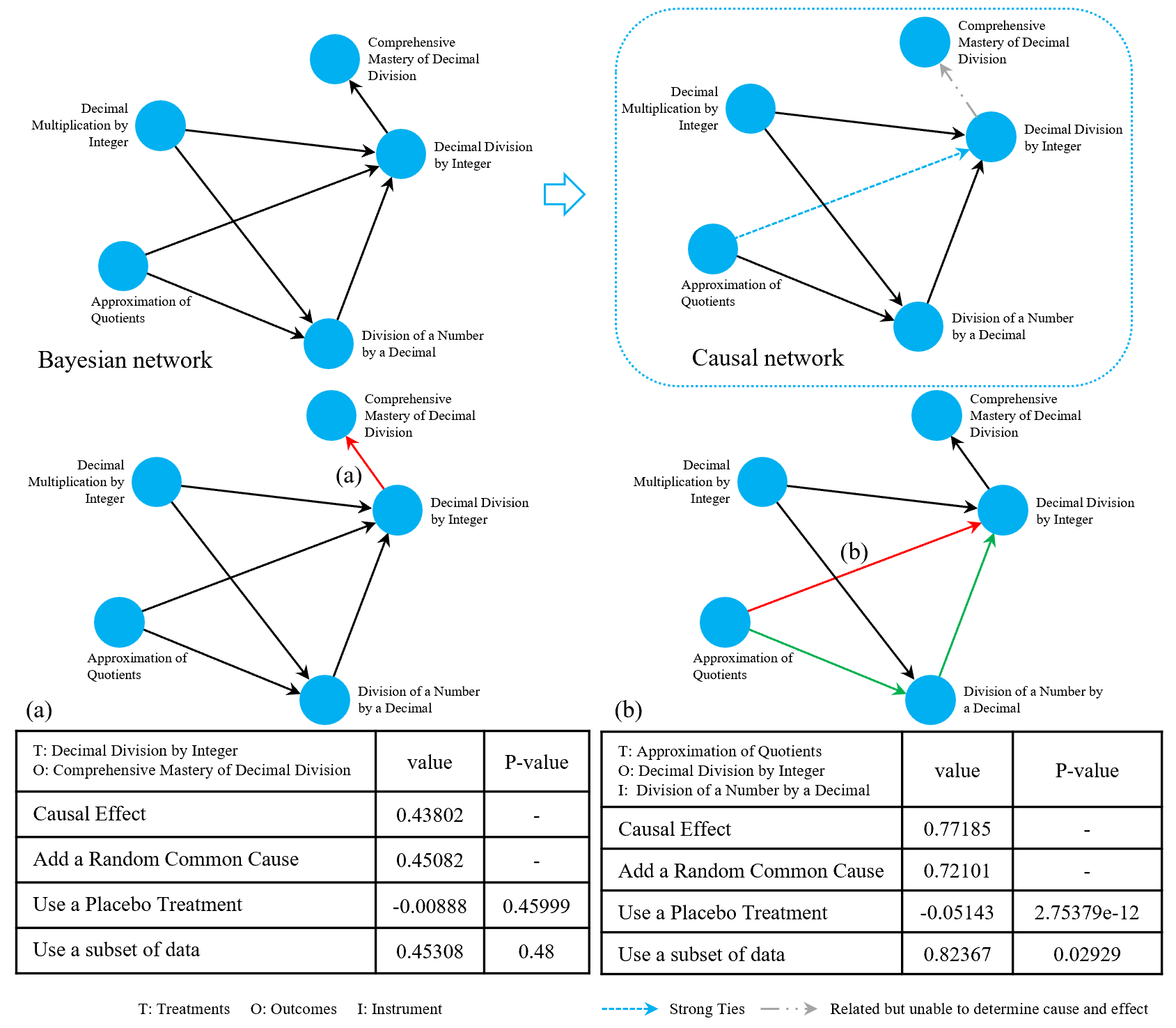}
\caption{Construction of knowledge component causal network}
\label{fig3}
\end{figure*}

As shown in Fig. \ref{fig3}, through the refutation experiment on edge (a), we found weak causal effect and low credibility of the correlation relationship, and therefore cannot admit the existence of causal relationships between nodes. In contrast, the refutation experiment on edge (b) revealed a causal effect of about 0.77 and over 95\% credibility, indicating a \textit{Strong tie} between the nodes. The modified network structure was then generated accordingly. Certainly, we can continue to explore the causal relationships between more nodes and update the network, in order to obtain the final causal network.

\section{Discussion}
A well-structured knowledge network can effectively support the development of ITS, enhancing personalization and improving educational quality. However, constructing such knowledge structures has always been a topic of interest, and the relationships between knowledge components can be challenging to elucidate. Most studies define knowledge structures through associative relationships, which do not necessarily represent the true causal links between knowledge components. 

During the construction of the causal network, we perform intervention and counterfactual experiments to discover and validate causal relationships. In the intervention process, we establish two new causal connections: \textit{Strong} and \textit{Weak tie}. Incorporating causal connection strength allows us to assess the nonlinear impact and differences in interaction strength\cite{rajkumar2022causal}. Counterfactual experiments further reinforce causal relationship judgments. These experiments construct a ``virtual" world to explore alternative potential outcomes, which helps make causal judgments.

Constructing causal knowledge component networks provides foundational insights for knowledge tracing (KT), learning resource recommendations, learning path planning, and learning outcome assessment. Specifically, utilizing feature causality can effectively select data features that enhance the performance of KT, while the causal relationships among these features can explain why a particular feature improves KT prediction outcomes\cite{jiang2024improving}. Moreover, discovering causal relationships among behaviors can elucidate which behaviors are causally linked to learning outcomes, providing teachers with actionable insights for instructional support\cite{jiang2023understanding}. Therefore, advancing research on knowledge component causal graphs can offer a foundational knowledge structure for building ITS. This structure can represent the relationships among knowledge components, aiding in the planning of students' learning sequences and the recommendation of practice resources based on the root causes of their issues. Additionally, feedback on the reliability and trustworthiness of the network structure from both teachers and students can be integrated with technical updates to continuously refine and enhance the knowledge component network, making it more accurate and explainable.

However, there are still some limitations and challenges; for example, the structural learning of large-scale Bayesian networks remains a challenging scientific problem, especially when analyzing causal relationships on this basis, which further increases the difficulty. The main challenges are high computational complexity, insufficient data, and uncertainty features. For instance, insufficient data in existing educational datasets, the largest online education dataset, EdNet\footnote[2]{https://github.com/riiid/ednet}, contains over 39,000 knowledge components, with approximately 15,000 in mathematics and 8,000 in science. On average, each knowledge component has only 341 data entries. This amount of data per knowledge component is insufficient for generating large-scale networks.

\section{Conclusion}
This paper focuses on constructing a causal knowledge network to investigate the causal relationships among learning concepts to enhance teaching quality and effectiveness. Using Bayesian networks and causal inference methods, a knowledge network based on learning performance data has been established, and these relationships have been validated and refined through intervention and counterfactual experiments. This work provides educators and learners with a comprehensive causal knowledge network that reflects the probabilistic relationships between concepts and provides insights into the causal mechanisms driving knowledge acquisition and mastery. Overall, this research offers a significant tool and approach for the education domain to promote explainability in educational technology and to provide personalized and effective learning support for learners.

\section{Acknowledgement}
This research was funded by the National Natural Science Foundation of China grant number 61977058, and the Natural Science Foundation of Shanghai grant number 23ZR1418500.

%
\bibliographystyle{abbrv}
%

\end{document}